IAC-19-D3.1.8

# Autonomous Multirobot Technologies for Mars Mining Base Construction and Operation

**Jekan Thangavelautham[a]\*, Aman Chandra[b], Erik Jensen[b]**

[a] Sp*ace and Terrestrial Robotics Exploration (SpaceTREx) Laboratory, Department of Aerospace and Mechanical Engineering, University of Arizona, 1130 N Mountain Avenue, Tucson, Arizona, 85721*, jekan@email.arizona.edu
[b] Sp*ace and Terrestrial Robotics Exploration (SpaceTREx) Laboratory, Department of Aerospace and Mechanical Engineering, University of Arizona, 1130 N Mountain Avenue, Tucson, Arizona, 85721*, achandra@email.arizona.edu
[c] Sp*ace and Terrestrial Robotics Exploration (SpaceTREx) Laboratory, Department of Aerospace and Mechanical Engineering, University of Arizona, 1130 N Mountain Avenue, Tucson, Arizona, 85721*, ejensen@email.arizona.edu

\* Corresponding Author

**Abstract**

Beyond space exploration, the next critical step towards living and working in space requires developing a space economy. One important challenge with this space-economy is ensuring the low-cost transport of raw materials from one gravity-well to another. The escape delta-v of 11.2 km/s from Earth makes this proposition very expensive. Transporting materials from the Moon takes 2.4 km/s and from Mars 5.0 km/s. Based on these factors, the Moon and Mars can become colonies to export material into this space economy. One critical question is what are the resources required to sustain a space economy? Water has been identified as a critical resource both to sustain human-life but also for use in propulsion, attitude-control, power, thermal storage and radiation protection systems. Water may be obtained off-world through In-Situ Resource Utilization (ISRU) in the course of human or robotic space exploration. Based upon these important findings, we developed an energy model to determine the feasibility of developing a mining base on Mars that mines and exports water (transports water on a Mars escape trajectory). Mars was selected as water has been found trapped in the regolith in the form hydrates throughout the surface at an average of 5% by mass with bigger deposits in Northern and Southern Polar Ice Caps. Our designs for a mining base utilize renewable energy sources namely photovoltaics and solar-thermal concentrators to provide power to construct the base, keep it operational and export the water using a mass driver (electrodynamic railgun).

Using the energy model developed, we determined that the base requires $3.1 \times 10^6$ MJ of energy per sol to export 100 tons of water into Mars escape velocity. 81.2% of the energy obtained from renewable power sources is to power the mass-driver. Only 17.9% of the energy is required to excavate, process and collect water. If the base was occupied by 100 human workers, 0.82% of the energy would be needed for sustaining life-support, food productions and healthy-living. Our studies found the key to keeping the mining base simple and effective is to make it robotic. Teams of robots (consisting of 100 infrastructure robots) would be used to construct the entire base using locally available resources and fully operate the base. This would decrease energy needs by 5-folds. Furthermore, the base can be built 5-times faster using robotics and 3D printing. This shows that automation and robotics is the key to making such a base technologically feasible.

**Keywords:** multirobot system, mining base, ISRU, Mars, excavation, human-robot teams.

## 1. Introduction

Beyond space exploration, the next critical step towards living and working in space requires developing a space economy. One important challenge faced with the space-economy is ensuring the low-cost transport of raw materials from one gravity-well to another. This is especially true of transporting material from Earth to off-world locations. The delta-v of 11.2 km/s makes this proposition very expensive. Transporting materials from the Moon takes 2.4 km/s and from Mars 5.0 km/s. Based on these factors, the Moon and Mars can end up becoming viable colonies to export material into this space economy. One critical question is what are the resources required to sustain a space economy?

Water has been identified as a critical resource both to sustain human-life, a near-universal chemical solvent but also for use in propulsion, attitude-control, power, thermal storage and radiation protection systems. Water is a critical resource because of its multi-functionality. Water may be obtained off-world through In-Situ Resource Utilization (ISRU) in the course of human or robotic space exploration that replace materials that would otherwise be shipped from Earth. Water has been highlighted by many in the space community as a credible solution for affordable/sustainable exploration. Water can be extracted from the Moon (though with some important uncertainties), C-class Near Earth Objects (NEOs), surface of Mars and Martian Moons Phobos and Deimos and from the surface of icy, rugged terrains of Ocean Worlds. United Launch Alliance, a collaboration between Boeing and Lockheed Martin is willing to buy water from any entity in space (in Low Earth Orbit, Geostationary Orbit and Lunar Orbit). In due course, we expect ULA would be willing to buy water in the Mars system as well. ULA has put forth a plans called "CisLunar 1000" [15] an architecture that foresees nearly 1000 people working and living cis-lunar space and the development of communication relays [13-14], service depots, and refuelling stations at strategic locations between Earth and Mars.

Based upon these important findings, we have developed an energy model to determine the feasibility of developing a base on Mars that mines and exports water (transports water on a Mars escape trajectory) so that a company such as ULA can buy it for its interplanetary transport needs. Mars was selected as water has been found trapped in the regolith in the form hydrates throughout the Martian surface at an average rate of 5% of the total mass. There are bigger deposits of $H_2O$-$CO_2$ in the Northern and Southern Polar Ice Caps. Our designs for a mining base (Fig. 1) utilize renewable sources of energy from the sun namely photovoltaics and solar-thermal concentrators to provide power to construct the base and keep it operational. This includes power to transport the water (export) at Mars escape velocities using a Mass Driver (electrodynamic railgun) powered using renewable energy. Such a transport device avoids the use of water as propellant and thus maximizes export of the mined water.

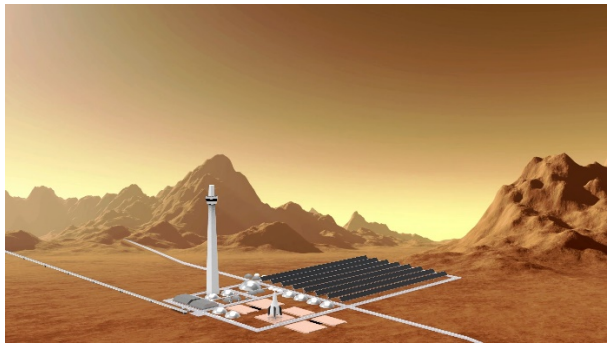

Fig. 1. Mars Mining Base Concept

Using the energy model developed, we determined that the base requires $3.1 \times 10^6$ MJ of energy per sol to export 100 tons of water into Mars escape velocity. 81.2% of the energy obtained from renewable power sources was to power the mass-driver to export the water into a Mars escape velocity. Only 17.9 % of the energy was required to excavate, process and prepare for export of the water. If the base was occupied by 100 human workers, another 0.82% of the energy would be needed for sustaining life-support, food productions and health-living.

Our studies found the key to keeping the mining base simple and efficient is make it a robotic base [1-6]. Teams of robots (consisting of 100 infrastructure robots with a mass of 120 kg each) would be used to construct the entire base using locally available resources and operate the base on a daily basis, mining water hydrates from the Martian regolith, refining it into liquid water and exporting it using a Mass Driver. Our studies found that a wholly robotic base built using 3D printing using raw natural resources can decrease energy needs by 5-folds. Furthermore, the base can be built nearly 5-times faster using robotics and 3D printing. If humans were in the loop, human energy needs overtake nearly-all other energy needs including excavation and processing resource material. This shows that automation and robotics is the key to making such a base technologically feasible. There are also other arguments why the base would benefit from being automated. This is because the tasks involved are 'dull', 'dangerous' and 'dirty' and so again this is ideal for robotics.

From these feasibility studies, we are now designing experimental prototypes of robots that would handle all the required tasks on a Mars Mining Base. In addition, we are designing experiments to demonstrate multiple-robots using renewable energy such as sunlight to perform 3D printing using sand and glass [9]. This includes sintering silica sand using sunlight in different proportions to produce everything from water-proofing/air-sealed glass, to bricks, columns and other critical building elements. According to our feasibility studies, these are some of the critical technologies required to construct and maintain the Mars Mining Base.

The proposed multirobot infrastructure technologies have major implications on Earth. Such robotics systems can be scaled up in terms of numbers to fully automate and make economically feasible whole new tasks not possible with current technology. This includes terraforming large tracts of desert into lush forests and grassland, turning hilly regions and mountain slopes ready for farming, dredging to form new islands and land in high-demand, highly populated cities. Furthermore, such robots maybe used to build and maintain sea-walls, water desalination systems and protect natural environments. Overall these technologies can be used to reduce the environment footprint of human civilization, making it carbon-neutral and minimizing impact on the plant and animal diversity on the planet

## 2. Related Work

Additive manufacturing (AM) is at the heart of the technology revolutionizing manufacturing and construction. The capability of AM techniques and the potential applications not yet realized fuel research and development into this emerging technology. In order to grasp what future potential AM possesses and where the current state of AM capabilities lie, the development history of this technology must be understood. The goal of this literature review is to gain a basic understanding of the wide range of AM techniques that currently exist and the applications for each technique.

Rapid prototyping (RP) was introduced to the world in the 1980's and has since morphed into what is commonly known as additive manufacturing, or 3D printing. The name rapid prototyping described the technology's original purpose of providing industry a cost effective and timely method for developing prototypes. The origin period offered two RP techniques: stereolithography (SL) and selective laser sintering (SLS) [8]. The original techniques were followed by Ballistic Particle Manufacturing (BPM), Laminated Object Manufacturing (LOM), Solid Ground Curing (SGC) and Three-Dimensional Printing (3DP). Despite increased competition in the RP market, all of the newly developed and patented techniques were aimed solely at industrial applications.

*2.1 Industry Applications*

Despite being conceived as a cost-effective and efficient method for industry to prototype products in development, advancements in technique, process and materials have allowed industry to utilize AM deeper in their production chain. The automotive and aerospace industries were on the ground floor when RP first emerged. Recognizing the potential benefits RP offered in the product development process, these industries have been involved with universities and research organizations working towards advancing the technology towards a viable manufacturing method. Non-critical parts produced by AM can be found on various aircraft. Aerospace companies like GE, Airbus, Rolls-Royce, BAE Systems and Boeing are currently employing AM for non-critical components and working towards the ability to manufacture critical components in the future. The automotive industry has employed RP and AM techniques in a similar fashion with future goals of producing replacement vehicle parts on demand instead of mass producing and storing them.

*2.2 Types of Additive Manufacturing Techniques*

AM techniques can be divided into five distinct categories of which we describe two of the most relevant for off-world applications. Some methods share similarities like materials used in production or the method for print control, but sharp differences allow for the categorical classification. There are also multiple methods that exist within each category. Table 1 presents the five techniques and highlights the individual characteristics of each. A more detailed description of Binder Jetting (BJ) and Material Extrusion (ME) follows.

*2.3 Binder Jetting (BJ): Overview and Process*

Binder Jet printing is a process of depositing an ink binding agent onto a thin powder bed layer [10]. The binding agent acts as a glue that fuses powder particles together into a solid material. The printer head contains a large amount of nozzles that dispense the binding ink to specified locations. BJ printing exist in both small and large scales. Industrial BJ systems offer larger build volumes than typical AM techniques (up to 2200 × 1200 × 600 mm). These large build volumes are capable of producing large scale objects, or multiple small-scale objects in a single print cycle. All binder material required to print a single layer is dispersed in a single pass of the printer head over the build area. Two different processes control the elevation of the print surface. In small BJ systems, the printer head and powder recoating blade remain at a constant height and the build platform lowers the distance of a single layer thickness to prepare for the powder recoat operation. In large scale BJ systems, it's common for the build platform to remain at a fixed elevation while the printer head and recoating blade incrementally rise between layers. The printer head and recoating blade are both supported by a gantry frame that encloses the entire build volume. Support structures are not needed in BJ printing because the entire build volume is filled with powder material. The unbounded powder serves a support to the printed object.

Binder Jetting is one of the few AM techniques that is being utilized to produce architectural scale objects. As can be imagined, a very large gantry frame is needed to enclose a large print area to make this possible. D-Shape$^{TM}$ is a company that has pioneered architectural BJ printing and now offers an architectural scale printer with a build volume of 12 × 12 × 10 m, shown in Fig. 2. D-Shape has been able to successfully produce structures like single story houses, small bridges and sculptures using their architectural scale BJ technology. Due to the volume constraints of the build volume, any structure that exceeds the size of the printer must be printed in modules and assembled. Another constraint of architectural scale BJ printing is the immobility of the printers. They are designed to be assembled and remain in a single location.

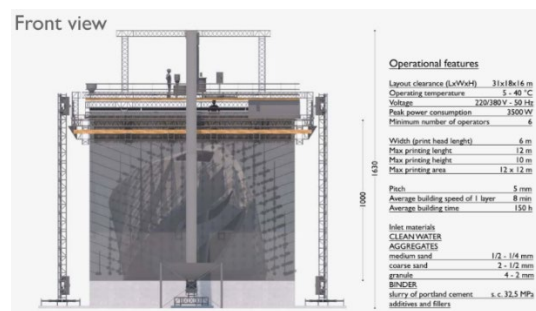

Fig. 2. D-Shape$^{TM}$ 31×18×16 m architectural scale binder jet printer.

Table 1. Additive Manufacturing Technology Comparison [10].

| Technique | Print Mechanism | Materials | Support Structure | Post-Processing | Mars/Moon |
|---|---|---|---|---|---|
| Vat Photo-polymerization (VP) | Directed Laser | Light activated polymers | Yes | Yes | No |
| Material Jetting (MJ) | Resin Dispensing Printhead (Gantry) | Thermoset photopolymer resin | Yes | Yes | No |
| Binder Jetting (BJ) | Binding agent dispensing printhead (Gantry) | Metals, sands, ceramics | No | Depends on application | Yes |
| Material Extrusion (ME) | Extrusion nozzle (Robotic or Gantry) | Wide range of thermo- plastics, cement paste, rubber, food paste | Possible | Depends on surface finish required | Yes |
| Powder Bed Fusion (PDF) | Laser or electron beam | Metals, Ceramics, Nylon | No | Depends on surface finish required | Yes |

*2.4 Post Processing*

All objects created using BJ require a cleaning process following printing. During the print process, the entire build volume is filled with the powder material being printed, but only a portion of that powder is used to create the desired object(s). The excess powder in the build volume must be removed. This is typically done by vacuum.

Objects printed using metallic powders require an additional process because they have weak mechanical properties straight out of the printer. The printing process yields an object that is simply metal particles bound together by a polymer binder. Infiltration and sintering are post processes that give the printed object desirable mechanical properties. Infiltration is done by placing the object in a furnace that
burns out the binding agent. At this point the object is approximately 60% porous.

The post processing of metallic objects introduces problems with final product dimensional accuracy. During infiltration, objects shrink by 2%. Sintering results in shrinkage of 20%.

*2.5 Binder Jetting Materials and Applications*

The most popular application of BJ printing is the production of sand molds used for metal casting. BJ allows for the creation of complex geometries that would otherwise be impossible with traditional mold production methods. These molds are comprised of sand or silica and are ready for use without the need for post processing. An example of a multi-part sand casting created by industrial BJ is shown in Fig. 3.

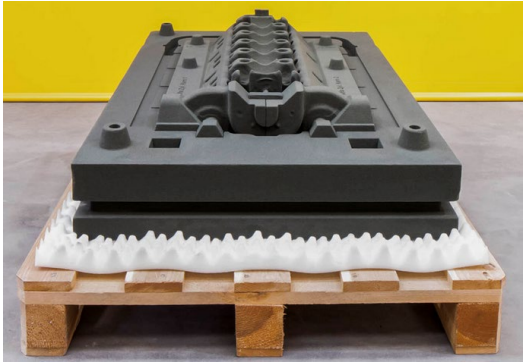

Fig. 3. ExOne® BJ printed sand mold for metal casting.

The metallic powders available to BJ printing are limited to only a few options. These include stainless steel, Inconel alloys and tungsten carbide. The production of metal parts using BJ are done at a fraction of the cost compared to other metal AM techniques. Despite the mechanical properties of metal objects produced by BJ not meeting requirements for high-end applications, parts produced are comparable to those produced by metal injection moulding.

*2.6 Material Extrusion (ME): Overview and Process*

Material extrusion is the most widely use d AM method at the consumer level. Material extrusion began in the 1980's under the name Fused Deposition Modeling$^{TM}$ (FDM) which has since been trademarked by Stratasys. Open source reference to material extrusion printing is referred to as Fused Filament Fabrication (FFF). Both names represent the same technique. The ME process generally uses a thermoplastic filament on spools. The wire-like material is fed through a heated extrusion nozzle that converts the solid thermoplastic material into a viscous liquid. The extrusion nozzle deposits the melted thermoplastic onto a build platform in a layered fashion similar to previously mentioned AM techniques. Initial versions of ME printers operated by the extrusion nozzle moving freely in the XY plane and the build platform lowering a distance equal to one-layer thickness after each layer was completed.

ME capability has been expanded to produce very large objects. Notable advancements in this area of printing can be attributed to a method invented by B. Khoshnevis at the University of Southern California [7]. The Contour Crafting Cooperation® resulted from the development of this technology has been licensed over 100 international patents by the University of Southern California. Contour Crafting (CC) applies material extrusion on an architectural scale by utilizing cement-based paste as its print material and controlling the extrusion nozzle with a gantry-like support structure. A representation of CC producing a building is shown in Fig. 4. The Contour Crafting Cooperation is also developing technologies for space applications that would be capable of building structures on other planets using on-site materials.

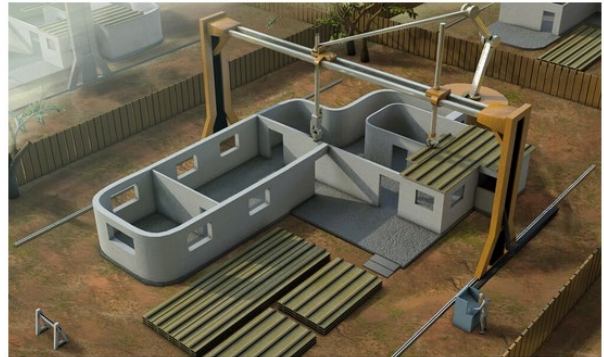

Fig. 4. Contour Crafting architectural scale printer [7].

Another pioneer in architectural scale ME is Apis Cor®. They have developed a mobile ME printing system comprised of a robotic printing arm and material supply that is capable of printing structures on site. The walls of a 400 m$^2$ house was successfully printed in 24 hours using this technology. Methods for printing foundations and roof structures are currently in development phases.

**3. Mars Mining Base Design**

The layout of the Mars Robotic Mining Base is shown in Fig. 1 and 5 and covers nearly 2 square kilometres. The mining base would be located at the base of a crater to exploit use of natural incline (slope) for the mass driver [11]. Key facilities on the base are interconnected by roadways constructed out of heat-fused silica (as a replacement to concrete). The command and control facilities of the base consist of a 150-meter high control tower to monitor/verify all operations on the base. The command tower will also be a localization beacon and tracking system for the fleet of autonomous robot. In addition, it is equipped with a communication ground station to communicate directly with Earth through the Deep Space Network (DSN).

Major facilities on the base including the refinery, service and repair centre, command and control buildings and even the warehouses are all 3D-printed spherical domes. The buildings are spherical domes to maximize internal volume with very minimal construction material. The domes would be constructed in one piece, include large glass windows, with an airtight fused silica glass layer in between to optionally maintain a 100 KPa, nitrogen/oxygen atmosphere inside.

A significant portion of the base is covered in solar-thermal and photovoltaic panels to harvest energy from the sun. Furthermore, solar-reflectors will be located on the crater rim to increase net-sunlight beamed to the base, in-addition to raising the temperature of the base surroundings. The base includes a hardened depot to house the 100 robotic vehicles when not in use, in-addition to service and repair facilities. The base houses a refinery to process (crush and bake to 120°C) the Martian regolith hydrate into water. An underground facility will be used to store both water and rocket propellant to refuel incoming rockets.

The human habitat and human facilities will be separated from the central operations area of the base by the Solar PV and solar-thermal generators. This is to minimize dust and sand being churned by moving vehicles entering and exiting the base from entering the human habitat regions. In addition, it will keep the human habitat sectors well away from the robotic vehicle traffic.

There are three modes of transport from the base. This includes roadways interlinking, (1) the base to the open pit mine site where the water-hydrate/ice deposits are located, (2) roadways reaching the rim of the crater and other parts of Mars and (3) a service road for the mass driver. The second mode of transport to and from the base is using rockets that can vertically land and take-off from one of six landing pads at the edge of the base. This provides quick access to the base in case of critical maintenance and for transporting repair equipment and other seed-resources to maintain continual operations of the base. Finally, the third form of transport is the mass-driver that will propel water housed in a container to 5.0 km/s (into a Mars escape trajectory for export).

In our simulation studies we consider four scenarios: Base constructed using (1) 3D printing of fused silica sand, with steel rebar support or (2) Steel structure with internal silica-sand blocks. Secondly, we consider the base to be (a) fully autonomous run using up to 100 mobile robots (Fig. 12) and (b) base run by a human team of 100 workers. The human occupied as would be expected is much more complex than the robotic base and includes additional buildings shaded in gray (Table 2). Importantly for the human occupied base, there will be six $O_2$ extractors that will electrolyze $CO_2$ in Martian atmosphere into breathable $O_2$. The base will also contain 2 large domes as large as the refinery housing services to take care of the human occupants, including health, banking, pharmaceutical, shopping, restaurants, rest and relaxation centre. The human occupants will live in individual dome-shaped housing with a minimal floor area 120 m² or 1,200 sq. feet. The housing areas will be linked via enclosed/pressurized above ground and underground walkways (tubes) spanning 10 km and connecting all the buildings. Several parallel tubes are in place to enable redundant and secure access to key facilities. In addition, there are parallel tunnels leading straight from the living quarters to the launch pad for quick evacuation in case of fire or a major onsite accident.

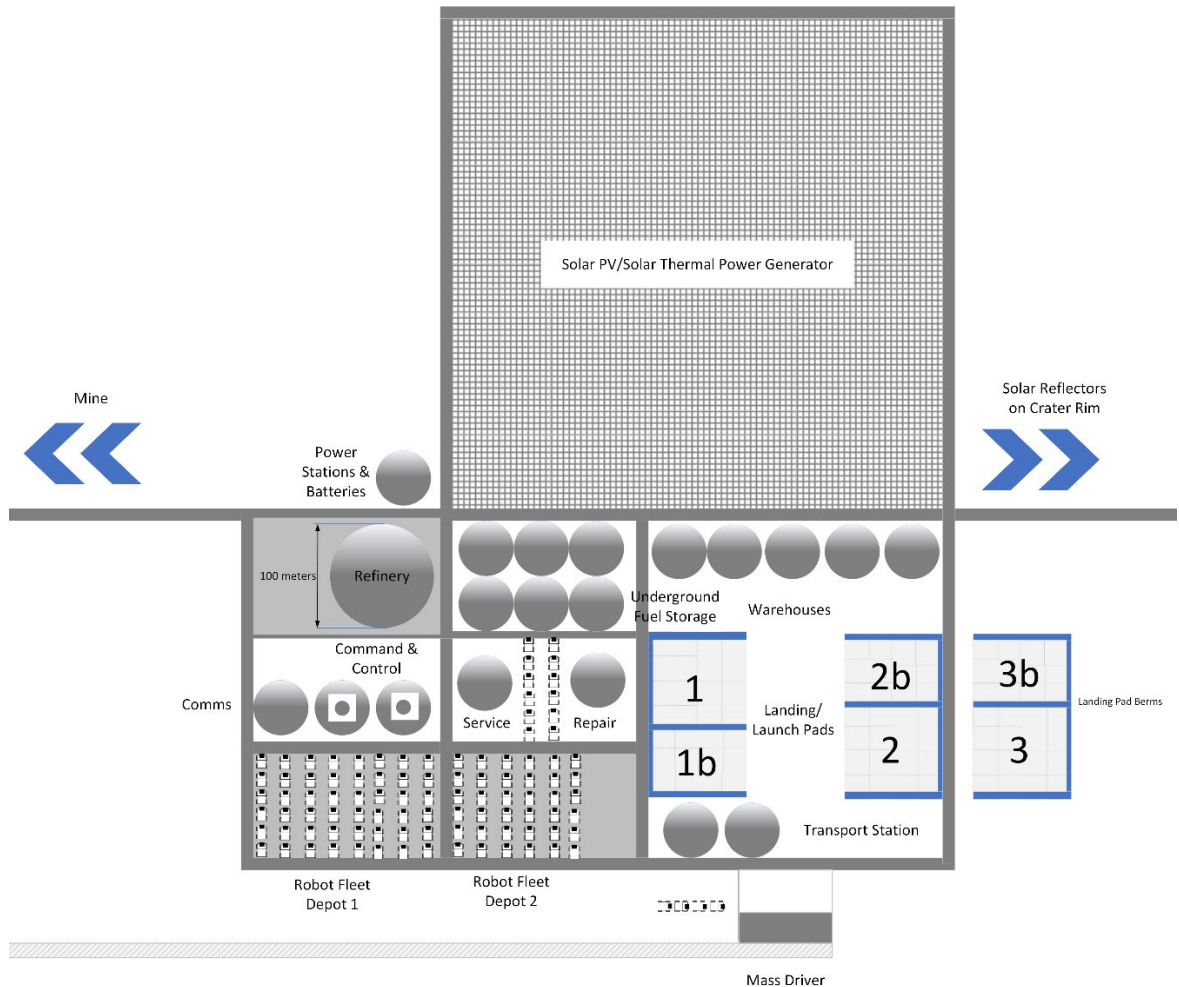

Fig. 5. Layout of a Mars Robotic Mining Base. It occupies over 2 sq. km and would be situated at the base of a crater with known water/hydrate/ice deposits. In this concept, there are no humans occupying the base.

The Mars Mining base contain the following infrastructure:

Table 2. Mars Mining Base Structures

| Structure | Quantity | Dimensions |
|---|---|---|
| Road Network [t] | 1 | Length, $L_t$ = 12,000 m, Width, $W_t$ = 8 m, Depth, $d_t$ = 0.2 m, Steel Ratio, $S_t$=0.05 |
| H₂O Extract Refinery [r] | 1 | Dome, Outer Radius, $R_{or}$ = 50 m, Inner Radius, $R_{ir}$ = 49.6 m, Base Depth, $D_r$=0.3 m, Steel Ratio, $S_r$=0.1 |
| Comms. Command, Control, Service and Repair, Power Ctrl [s] | 6 | Dome, Outer Radius, $R_{os}$ = 25 m, Inner Radius, $R_{is}$ = 24.6 m, Base Depth, $D_s$=0.3 m, Steel Ratio, $S_s$=0.1 |
| Control Tower [CC] | 1 | Pyramidal Tower, Height, $H_{cc}$=150, Base Length, $L1_{cc}$=25, Top Length, $L2_{cc}$=5, Steel Ratio, $S_{cc}$= 0.1 |
| Warehouses [w] | 6 | Dome, Outer Radius, $R_{ow}$ = 25 m, Inner Radius, $R_{iw}$ = 24.8 m, Base Depth, $D_w$=0.3 m, Steel Ratio, $S_w$=0.1 |
| Fuel Storage [f] | 6 | Cylinder, Radius, $r_f$ = 25 m, Wall Thickness, $D_f$=0.1 m, Wall Height, $h_f = 2\ m$, Steel Ratio, $S_f$=0.05 |
| Small Landing Pad [p] | 3 | Pad and Blast Walls, Length, $L_p$ = 100 m, Width, $W_p$ = 50 m, Depth, $D_p$ = 0.2 m, Wall Height, $H_p$ 0.5 m, Steel Ratio, $S_p$=0.05 |
| Large Landing Pad [p] | 3 | Pad and Blast Walls, Length, $L_p$ = 100 m, Width, $W_p$ = 100 m, Depth, $D_p$ = 0.2 m, Wall Height, $H_p$=0.5 m, Steel Ratio, $S_p$=0.05 |
| Power Generation Pad [g] | 1 | Pad, Length, $L_g$=1000 m, Width, $W_g$= 1000 m, Depth, $D_g$=0.1 m, Steel Ratio, $S_g$=0.05 |
| Mass Driver [m] | 1 | Cylinder (Pair) + Support, Inner Radius, $R_m$ = 3 m, Length, $L_m$ = 10,000 m, Thickness, $T_m$ = 0.5 m, Slope, $\alpha_m$ = 5°, Post Height= $H_m$= 5 m, Post Width = $W_m$ = 1 m, Post Spacing = $\varepsilon_m$ = 20 m, Post Steel Ratio, $S_m$=0.05 |
| O₂ from CO₂ atm. Extract* [O] | 6 | Cylinder, Radius, $r_o$ = 50 m, Wall Thickness, $D_o$=0.2 m, Wall Height, $h_o$=3.0 m, Steel Ratio, $S_o$=0.2 |
| Human Services Centres* [SC] | 2 | Dome, Outer Radius, $R_{or}$ = 50 m, Inner Radius, $R_{ir}$ = 49.6 m, Base Depth, $D_r$=0.3 m, Steel Ratio, $S_r$=0.1 |
| Human Habitat* [h] | 100 | Dome, Outer Radius, $R_{oh}$ = 6 m, Inner Radius, $R_{ih}$ = 5.9 m, Base Depth, $D_h$=0.1 m, Steel Ratio, $S_h$=0.2 |
| Human Walkways (Service Tubes) [ST] | 1 | Cylinders, Length, $L_{ST}$ = 10,000 m, Outer Radius, $R_{oST}$ = 1.1 m, Inner Radius, $R_{iST}$ = 1.0 m, Steel Ratio, $S_{ST}$=0.1 |

The equations for volume calculation of the construction material for the base structures is given below. Table 2 shows the values of the variables used and subscripts.

$$V_t = L_t W_t D_t \quad (1)$$

$$V_r = \tfrac{2}{3}\pi\left(R_{or}^{\ 3} - R_{ir}^{\ 3}\right) + \pi D_r R_{or}^{\ 2} \quad (2)$$

$$V_s = \tfrac{2}{3}\pi\left(R_{os}^{\ 3} - R_{is}^{\ 3}\right) + \pi D_s R_{os}^{\ 2} \quad (3)$$

$$V_{CC} = H_{cc} L_{2cc}^2 + H_{cc}(L_{1CC} - L_{2CC})L_{2CC} \quad (4)$$

$$V_w = \frac{2}{3}\pi(R_{ow}^3 - R_{iw}^3) + \pi D_w R_{ow}^2 \tag{5}$$

$$V_f = 2\pi r_f^2 D_f + 2\pi r_f h_f D_f \tag{6}$$

$$V_p = L_p W_p D_p + 2L_p H_p D_p + W_p H_p D_p \tag{7}$$

$$V_g = L_g W_g D_g \tag{8}$$

$$V_O = 2\pi r_o^2 D_O + 2\pi r_o h_o D_O \tag{9}$$

$$V_{SC} = \frac{2}{3}\pi(R_{oSC}^3 - R_{iSC}^3) + \pi D_{SC} R_{oSC}^2 \tag{10}$$

$$V_h = \frac{2}{3}\pi(R_{oh}^3 - R_{ih}^3) + \pi D_h R_{oh}^2 \tag{11}$$

$$V_m = 2\pi L_m (R_m + T_m)^2 - 2\pi L_m (R_m)^2 + \frac{L_m}{\varepsilon_m} 2H_m W_m^2 \tag{12}$$

$$V_{ST} = \pi(R_{oST}^2 - R_{iST}^2)L_{ST} \tag{13}$$

*3.1 Energy for Silica Sand 3D Printing + Reinforcement*

The energy needed for silica sand 3D printing involves first raising the temperature of the sand to $\Delta T_{sand}$ =1,973 °K followed by melting the sand into a liquid binder. The heat capacity of sand (quartz) is, $C_{p\ sand}$= 0.830 kJ/(kg K). The heat of fusion to melt sand (quartz) is $H_{melt}$=156 kJ/kg. Therefore, the total energy needed to melt and fuse the sand insitu is the following

$$E_{total\ sand} = E_{heat\ sand} + E_{melt\ sand} + E_{trans\ sand} \tag{14}$$

$$E_{total\ sand} = \rho_{sand} V (C_{p\ sand} \Delta T + H_{melt\ sand} + gd) \tag{15}$$

The density of quartz sand is $\rho_{sand} = 1500\ kg/m^3$ and g is the local acceleration due to gravity and d is distance in metres. On Mars g=3.71 m/s². Unless if the sand is transported long distances in the order of 1000s of metres, the energy required simplifies to the following:

$$E_{total\ sand} \approx E_{heat\ sand} + E_{melt\ sand} \tag{16}$$

For steel-bar reinforced structures, the total energy is the following:

$$E_{total\ reinf} = E_{total\ sand} + E_{total\ steel} \tag{17}$$

$$E_{total\ reinf} = \rho_{sand}(1-S)V(C_{p\ sand}\Delta T_{sand} + H_{m\ sand}) + SV\rho_{steel}(C_{p\ steel}\Delta T_{steel} + H_{m\ steel}) \tag{18}$$

Where S is the volumetric percentage of steel, $\rho_{steel}$ = 7,750 $kg/m^3$, $C_{p\ steel}$ = 0.510 kJ/(kg K) and $H_{m\ steel}$ = 25.23 MJ/kg where the steel is produced from Martian regolith containing 1:1, Magnetite and Hematite with heat of melting 1118 kJ/mol and 824 kJ/mol. Only 70% of Magnetite and Hematite contain Iron, with the remainder being oxygen. $\Delta T_{steel}$ =1,640 °K

*3.2 Steel and Sand Block Construction*

For the steel and sand-block construction, the structures are all made of a steel support structure and thus the volume of the structure is µ=0.15 of the 3D printed structure. For road, the volume of the structure is µ=0.05 of the 3D printed structure. Similarly, to form the sand blocks, only µ percentage volume of the sand is melted thus energy equation is the following:

$$E_{total} = \rho_{steel} V \mu (C_{p\ steel}\Delta T + H_{m\ steel}) + \rho_{sand} V \mu (C_{p\ sand}\Delta T + H_{m\ sand}) \tag{19}$$

*3.3 Water Extraction Energy*

We wish to extract water from the Martian regolith contained in the form of hydrates. We model the scenario as the regolith contain 90.6 % sand (quartz) and 9.4 % $MgCl_2 \cdot 6H_2O$. This gives us ~5 % $H_2O$ by mass.

To extract water from m= 100,000 kg, from $MgCl_2 \cdot 6H_2O$ requires raising the temperature to 111 °C from a presumed ambient of 25 °C, where the heat capacity $C_{pmg}$= 0.756 kJ/(kg K) The heat of dehydration is the $E_{dehyd}$= 138 kJ/kg. Furthermore, the extracted water needs to be condensed down to 25 °C. Therefore, the total energy required is the following:

$$E_{heat} = E_{dhyd}m(9.4/5) + (111\text{-}25)[2C_{pH2O}m + E_{cndH2O}m + C_{pmg}(4.42/5)\ m + C_{psand}m(90.6/5)] \tag{20}$$

Based on this estimate, it takes $E_{heat}$ = 4.37 × 10⁵ MJ to extract the water from the regolith.

*3.4 Excavation and Regolith Transport Energy Req.*

We wish to calculate the total excavation energy required to produce 100,000 kg of water a Martian sol. First, we calculate the transport energy required to move Martian regolith from open pit mining site to refinery with 5% water by mass a distance d = 2000 m, with friction coefficient η=0.05, χ= 1.2 and is the vehicle movement ratio.

$$E_{mov\ up} = (100/5)\ mg\eta d\chi \tag{21}$$

Plugging in the values, $E_{mov\_unp}$ = 864 MJ. Next calculated the total transport energy required to transport the processed water from refinery to mass driver. We presume a maximum d= 1000 m, with friction coefficient η=0.01:

$$E_{mov\ p} = mg\eta d. \qquad (22)$$

Next, we calculate an estimate of the total excavation distance required to cover a flat open pit. Knowing that the $\rho_{sand}$ = 1500 kg/m³, we determine the total volume of the regolith. Following this we divide by the total area of the robot vehicle, $A_{robot}$ = 0.42 m² and account for picking up and returning with regolith. The expression is then the following:

$$E_{dig\ p} = [\chi F_{dig} + (100/5)mg\eta]\ [(100/5) \qquad (23)$$
$$m/\rho_{sand}]\frac{2}{A_{robot}}$$

Where χ= 1.2 is vehicle movement ratio and accounts for inefficiencies in movement. $F_{dig}$ = 3000 N is the force used to dig into the Martian regolith and m= 100,000 kg and η=0.1 is the friction coefficient and is applicable for movement in rough terrain. Then $E_{dig\ p}$ = 4594 MJ. With this the total energy required for excavation, transport of unprocessed regolith, water extraction and transport of water to mass driver is $E_{Exc\_transp}$ = 5.51× 10⁵ MJ.

*3.5 Mass Driver Energy Needs*

For the proposed mining base, a Mass Driver will be used to send the m=100,000 kg of liquid water in a container consisting of M=10,000 kg into a v=5000 m/s Mars escape velocity and d= 40,000 m, $C_d$=0.01, A= 7.06 m², ρ=0.02 kg/m³. The total energy requirement is the following:

$$E_m = \frac{1}{2}(m+M)v^2 + \frac{1}{2}\rho C_d A v^2 d \qquad (24)$$

To overcome any drag close to the atmosphere and assuming drag going as far as 40 km altitude. We simply increase the overall kinetic energy to account for drag losses.

$$E_m = \frac{1}{2}(m+M)v_*^2 = \frac{1}{2}(m+M)v^2 + \qquad (25)$$
$$\frac{1}{2}\rho C_d A v^2 d$$

$$v_* = \left(v^2 + \frac{\rho C_d A v^2 d}{m+M}\right)^{1/2} \qquad (26)$$

$$v_* \approx 5001.8\ m/s \qquad (27)$$

The total kinetic energy required is then $E_m$=2.5 × 10⁶ MJ

*3.6 Energy Requirements for Work Crew*

Here is a table of daily energy requirements for the 100-member work crew:

Table 3. Life Support Energy Consumption

| Item | Energy Required |
|---|---|
| Oxygen Generation | 5,559 MJ |
| Electricity | 5,400 MJ |
| Water Needs | 10,800 MJ |
| Food | 3,354 MJ |
| Total | 25,113 MJ |

These figures are based on expected requirements for an average adult human. An average adult is expected to breathe 98.2 moles of oxygen a day. The energy required to make Oxygen on Mars is 566 kJ/mol. Each person is estimated to consume 15 kWh per sol for personal use which is 5,400 MJ for 100 workers. Each worker is assumed to consume 300 L in water a day. The water is recycled and presumed to consume 1 kWh/L. In terms of food, the following Table 4 shows kJ energy consumed in terms of food person. Considering 50% of food goes to waste, we presume the total requirements of food per person sol is 16.77 ×2 kJ = 33.54 kJ.

Table 4. Human Daily Food Consumption

| Food | Energy Used [kJ/kg] | Consumed [kg] | Total Energy [kJ] |
|---|---|---|---|
| Corn | 1.1 | 0.306 | 0.336 |
| Milk | 2.2 | 0.18 | 0.39 |
| Fruits & Vegetables | 4.4 | 0.864 | 3.80 |
| Eggs | 8.36 | 0.09 | 0.75 |
| Chicken | 8.8 | 0.09 | 0.79 |
| Cheese | 17.6 | 0.09 | 1.58 |
| Goat | 30.8 | 0.09 | 2.77 |
| Beef | 70.4 | 0.09 | 6.33 |
|  |  | 1.8 | 16.77 |

Based on these calculations, the total energy requirements to support the 100 human workers is 25,113 MJ/sol.

*3.7 Energy Harvested for the Mining Base*

The presented base will be using renewable energy to power the entire facility. The base needs to generate 3.1 × 10⁶ MJ/sol which is a combination of electrical and thermal energy. For this we make some simplifying assumptions. The average solar insolation on Mars is taken to 40% of that of Earth which is 1350 W/m². This provides an average of 540 W/m². Solar-thermal systems that use Carbon Nano-Tubes (CNTs) can directly convert solar energy into heat at 99% efficiency. Using CNT, we consider assembly of a solar-thermal plant that capture solar heat. We presume we require generating $3.1 \times 10^6$ MJ/sol of thermal power and furthermore sunlight hours last 7 hours/sol and the average solar insolation, χ=540 W/m². The total size of the solar-thermal power plant requires is the following:

$$E_{solar} = \Delta T \chi A_{solar} \quad (28)$$

This requires a total area of 0.226 km². Next, we presume the total energy required per day is all electricity. Then equation for the total energy required is the following:

$$E_{PV} = \Delta T \chi \lambda_{PV} A_{PV} \quad (29)$$

Where $\lambda_{PV}$ is the photovoltaic efficiency. We presume it is 45%. With these parameters, the required Area, $A_{PV}$ is 0.50 km².

## 4. Results and Discussion

Autonomous robotics and 3D printing can have game-changing impact on Mars Mining Base construction and operation. We developed an energy model that account for construction, operation and maintenance of a Mars Mining Base. In this study, we also considered the use of 100 human workers vs. 100 infrastructure robots to operate and maintain the base.

Use of robots and 3D printing can reduce energy required for base construction by 5-folds and it can speed-up construction of this mining base by 5-folds. It can simplify base designs, leading to greater robustness and more rapid construction. Humans consume more energy for their daily needs than total energy required to excavate, process and ground transport the resources. This shows that automation and robotics is the key to making such a base technologically feasible. Here we analyse the feasibility of developing a Mars Robotic Mining Base with the emphasis on mining water-ice into rocket fuel. The base will export 100 tons of water ice per sol (1 Martian day, 24.6 hours).

A significant portion of the base is covered in solar-thermal and photovoltaic panels to harvest energy from the sun (see Section 3.7). Furthermore, solar-reflectors will be located on the crater rim to increase net-sunlight beamed to the base, in-addition to raising the temperature of the base surroundings. The base includes a hardened depot to house the 100 robotic vehicles when not in use, in-addition to service and repair facilities. The base houses a refinery to process (crush and bake to 120º C) the Martian regolith hydrate into water. An underground facility will be used to store both water and rocket propellant to refuel incoming rockets.

This design of a Mars Robotic Mining Base is quite modular, with all of the buildings and structures in pre-fab configuration thus facilitating 3D printing of parts. The base can be steadily expanded from each side, adding for example one or more refineries, robot fleet depots, solar-thermal and solar-photovoltaic generators to enable increased water export.

*4.1 Energy Required for Construction*

Our studies compared the potential options for building a Mars Mining Base (Fig. 6). This included use of humans or robots and 3D-printing or no 3D printing. Our studies find that 3D printing and robotics provides the biggest advantage. It can cut time to build by 5 folds and decrease energy consumption by 5 folds (Fig. 6,7).

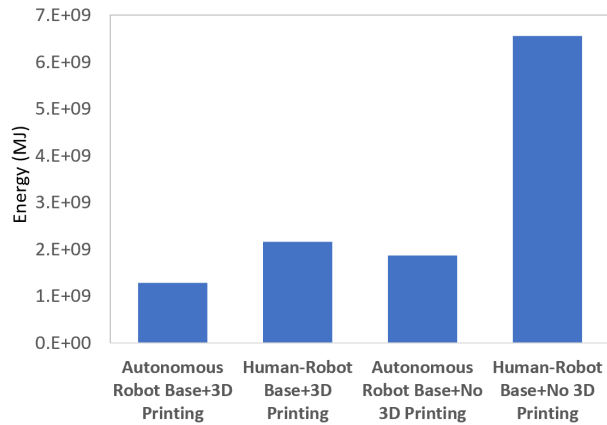

Fig. 6. Building a Mars Mining Base using a human team and without 3D printing requires 5-folds more energy than without. This show robotic 3D-printing has significant potential to lower cost and make feasible a Mars Mining base.

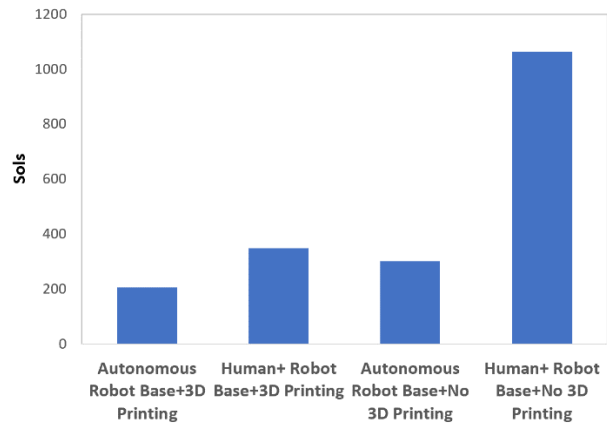

Fig. 7. Our model shows that with the available renewable power source it will take 5 folds longer to build a base using a human team without 3D printing than using a wholly robotic team 3D printing.

Based on our model, the significant simplifications possible of using an all-robot-team and housing robots is significantly more cost-effective than having humans and providing the significant resources needed to enable healthy living (Fig. 8, 9). If the resources needed for the base are obtained locally, that makes significant difference in terms of required transport energy and cost. Hence this study shows the potential game-changing opportunity possible with a construction method that utilizes local resources and doesn't require humans in the loop.

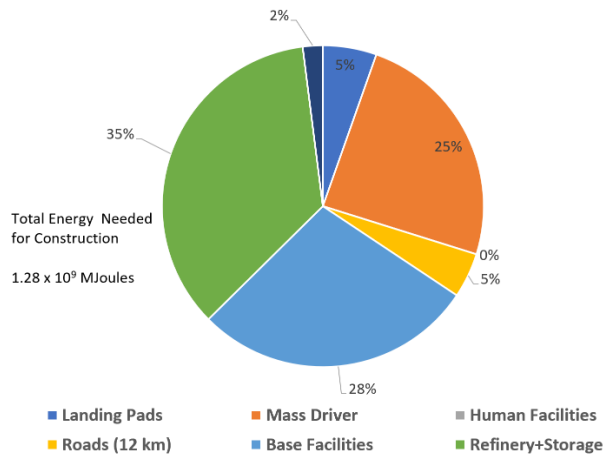

Fig. 8. Distribution of energy consumed for building a robotic mining base using 3D printing.

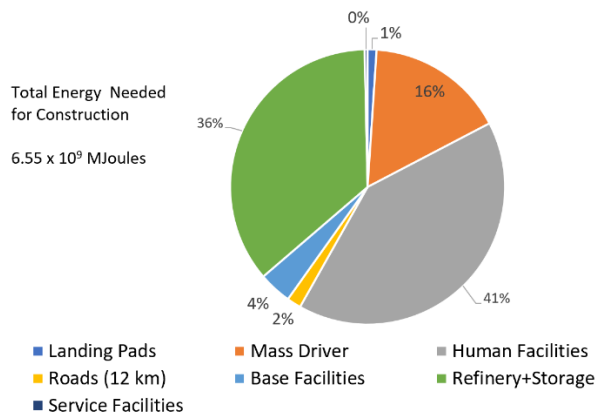

Fig. 9. Distribution of energy consumed for building human occupied mining base using conventional construction methods (no 3D printing).

We look into the details of building the Mars Robotic Mining Base (Fig. 8, 9). We presume 12 km of paved roads being built by fused silica-sand with properties similar to concrete. The roads will be reinforced with steel support beams to maximize strength for a 100-year lifetime. These roads will form the transport network for the fleet of 100 robot vehicles which is shown in Section 4.4. Importantly the roads will connect the open-pit mine to the base and reflector facilities on the crater rim. The remainder will be base facilities including the power station, communication ground station and command and control facilities and 150-meter control tower.

The building structures are all domes that are 3D printed in place using the local silica sand and will be held with a reinforced steel structure that would also be locally produced from the Martian iron-rich regolith. The refinery will be the biggest dome structure on the base with a diameter of 100 m and hardened further to prevent any spill-over from damage due to accidents. The remaining 17% of the total energy will be spent on building the robot facilities including depot, maintenance and repair facilities, together with parts storage.

*4.2 Energy Required for Base Maintenance*

According to our model, the total energy required for maintaining a Mars Mining base is $3.1 \times 10^6$ MJ per sol. Of this, 81.3% of the required energy is for transporting the 100 tons of water into Mars escape trajectory (Fig. 10). Remaining 17.9 % would be for mining, processing and maintaining the base facilities, while 0.82 % would be the energy required if there were to be 100 human workers. The significant amount of energy needed to export the water from Mars is unavoidable due to the Martian gravity. Building a Mass Driver [11] to transport this quantity of water will be much easier than to do so from Earth. This will also be significant savings in terms of the water being mined.

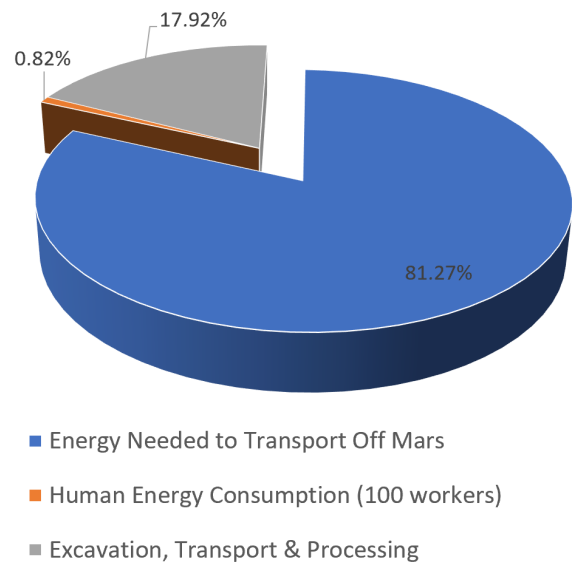

Fig. 10. Distribution of energy consumed for operating a humans-occupied Mars Mining base.

*4.3 Human Impact*

Based on our models, the energy consumed per sol (Martian day) for operation of the base (excluding Mass Driver transport) is $5.7 \times 10^5$ MJ. Our studies show that having humans on a Martian mining base results in significant consumption of energy and facility needs (Fig. 8-11).

In fact, quite rapidly, human energy needs overtake all other energy needs except mass driver transport and heating of regolith to extract water. This in part because Mars in its current form is an unfavourable environment to support Earth life and hence significant increase in energy and resources are required to sustain the life of the 100 human workers. In our energy model, we breakdown human needs into water, oxygen, food and electricity. It is presumed each worker will require about 300 litres of water per day. The highest per capita usage of water is 550 litre per day in the United States. We presume 300 litres as it is attainable and will be possible with improved efficiency expected on a future Martian base.

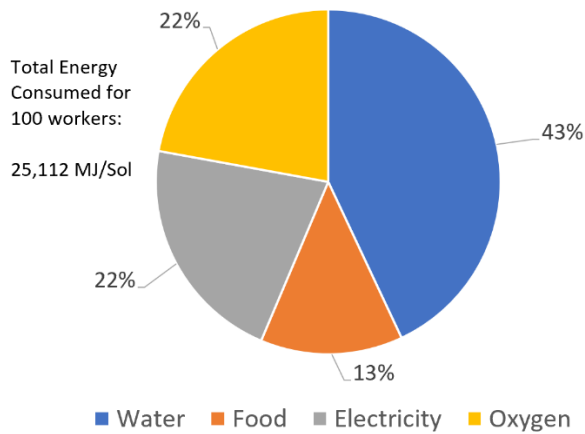

Fig. 11. Distribution of energy consumed by 100 human workers on the Mars Mining Base
.

In addition, each worker is assumed to be consuming 15 kWh of electricity per day and 2,200 litres of oxygen. In terms of food (Fig. 11), each worker is estimated to consume nearly 1.8 kg of food (2,800 calories) per day, with 35% of the food being fats and proteins (including milk, eggs, meats and cheese), 17% being starch (wheat, corn, rice) and 48% (fruits, vegetables). We then accounted for net energy required to produce these foods. Beef is estimated to require 70.4 kJ/kg, while corn for example requires 1.1 kJ/kg. Using this detailed model, we find that the 100 workers consume 25,113 MJ/sol.

This shows that automation and robotics is the key to making such a base technologically feasible. In the next section we provide a summary of our plans to further develop the 3D printing technology.

*4.4 Robotics Vehicles for 3D Printing, Excavation and Maintenance*

The proposed robotic vehicles collect regolith and processes them onboard to perform 3D printing (Fig. 12, 13, 14). The size of the 3D printed object is not limited by the size of the vehicle. The vehicles are powered entirely on renewable energy, using high-energy fuel cells that provide double the energy output of gasoline. The robotic vehicles will also be autonomous operating as a group, with only high-level commands being provided by a human supervisor from Earth or a relay base [1-6] (see Fig. 15). Teams of autonomous vehicles under the right conditions can exceed human controllers.

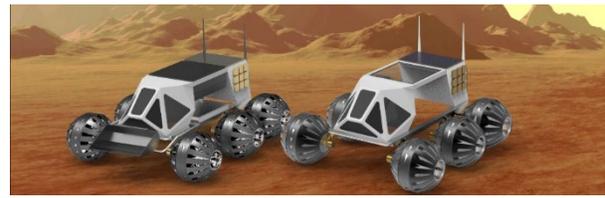

Fig. 12. 3D Rendering of a pair of robotics vehicles for excavation and 3D printing.

The vehicles are entirely autonomous and utilizes a combination of onboard sensors, GPS and local beacons to navigate. These vehicles are the template for a family of infrastructure robotic vehicles. On Mars they can be used on everything from the trenching of underground conduits for water, air, power and data, to the construction and maintenance of pressurized greenhouses. They can be used to build roads, bridges, and more advanced infrastructures for human settlement including radiation-shielded architectures. Gravity being 38% that of Earth, smaller units can be effective at moving large volumes.

On Earth they can be used to dig and maintain canals, tend to soil and plants on a massive scale (reclaiming land and minimizing the use of pesticides and herbicides), identify and harvest ripe crops, control invasive plants, and develop forest canopies.

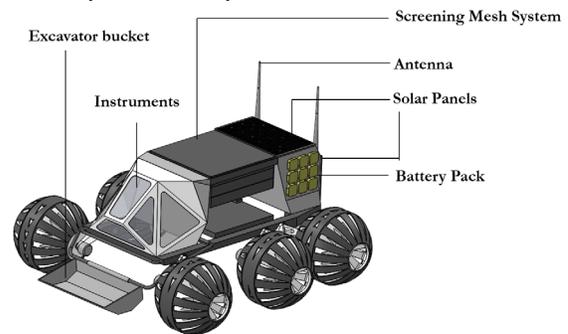

Fig. 13. 3D Rendering of an autonomous robotic vehicle for excavation.

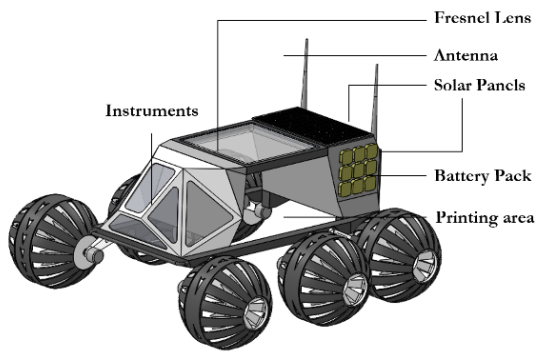

Fig. 14. 3D Rendering of an autonomous robotic vehicle for 3D printing.

Here we present advanced laboratory prototypes of the vehicles that we expect to build to test end-to-end automated design, excavation and 3D printing capability in a controlled setting on Earth as part of the next phase of this research program (Fig. 16, 17).

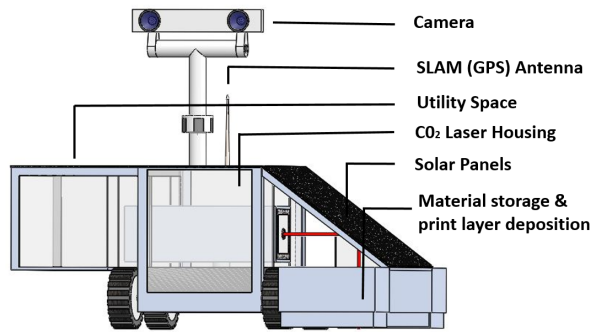

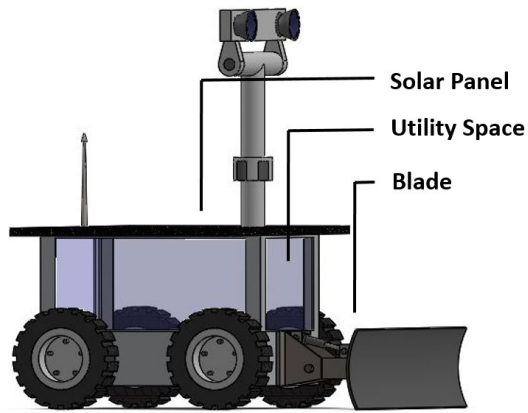

Fig. 16. An advanced lab prototype of a 3D Printing Rover (top) and Excavating Rover (bottom).

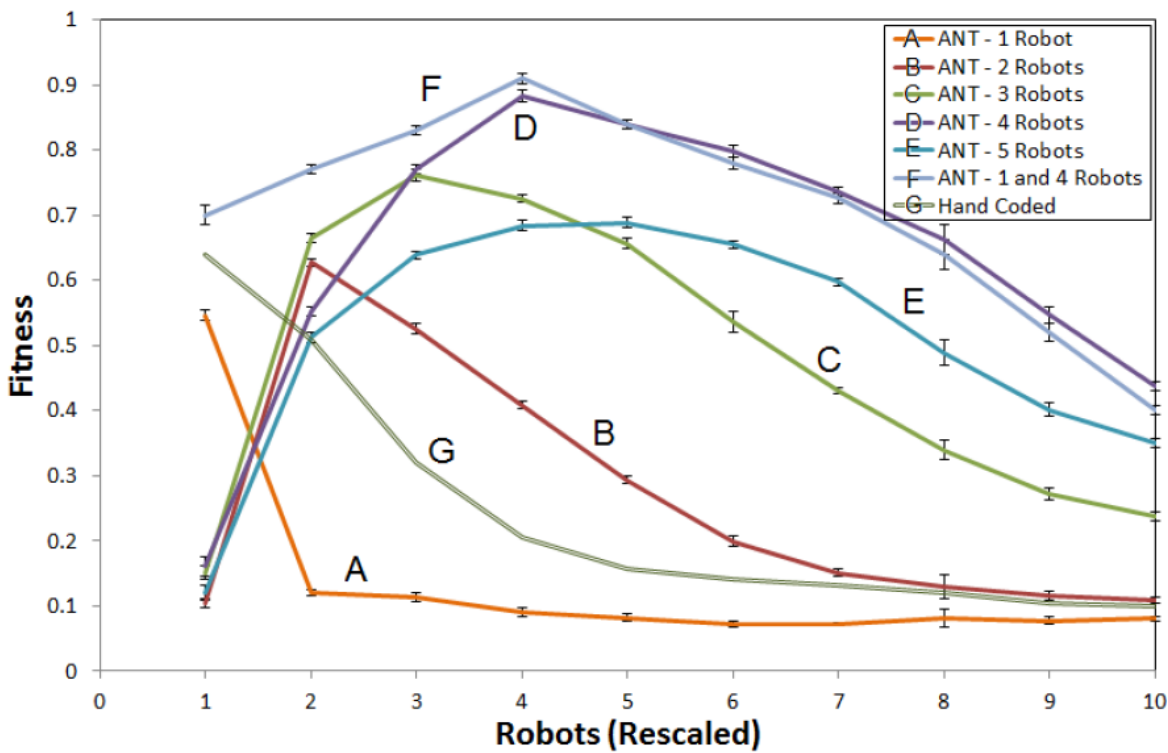

Fig. 15. Our earlier work shows that teams of robots for excavating task can be more effectively controlled using autonomous decentralized neuromorphic controllers than human-coded program [2, 12].

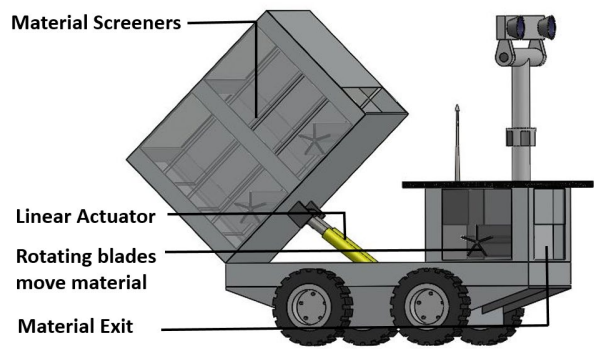

Fig. 17. Regolith Processing Rover

## 5. Conclusions

In this feasibility study we analyzed the application of multirobot systems towards development of a Mars Robotic Mining base. We identified mining on Mars to be feasible due to the abundance of water hydrates that could be crushed and baked to obtain liquid water. Plans are already underway in developing a space economy, with water being exchanged and traded at strategic locations. Water is critical to this space economy as it can be electrolyzed into hydrogen and oxygen to power a high-performance interplanetary propulsion system. Having to not carry the required fuel from Earth has a multiplier effect on all future space missions, as they can be low-cost, low-mass and be less complex. Such a system will also be sustainable enabling end-to-end reuse of rockets.

Our studies show that a precursor to a Mars or Moon Robotic Mining Base is technologically feasible within the next 5-10 years. The critical technologies, namely autonomous multirobot systems [1-6] and 3D printing have undergone significant advancement and are being prepared for rugged-field use. The power of this technology lies in removing humans out of the loop. This reduces cost of labor and enables a whole range of automation tasks to be handled by robots. Using control system algorithms developed in our lab [1-6], we have methods to scale up the technology to hundreds if not thousands of robots. The potential for this technology is not just cost-effective infrastructure development off-world but it can be readily be extended Earth. The technology is powered by renewable energy and can be scaled up based on task demands. Such a technology can tackle what are otherwise uneconomic, labor intensive tasks using present-day conventional technology. This includes turning deserts into green grassland and lush forest, provide fresh water canals systems to arid regions, restructuring hills and mountain slopes for agriculture and for dredging and forming new island and extended coast-lines in cities and populated areas.

## Acknowledgements

This project received seed funding from the Dubai Future Foundation through Guaana.com open research platform.